# What shall we do with an hour of data? Speech recognition for the un- and under-served languages of Common Voice

**Francis M. Tyers**
Department of Linguistics
Indiana University
ftyers@iu.edu

**Josh Meyer**
Coqui
josh@coqui.ai

## Abstract

This technical report describes the methods and results of a three-week sprint to produce deployable speech recognition models for 31 under-served languages of the Common Voice project. We outline the preprocessing steps, hyperparameter selection, and resulting accuracy on official testing sets. In addition to this we evaluate the models on multiple tasks: closed-vocabulary speech recognition, pre-transcription, forced alignment, and key-word spotting. The following experiments use Coqui STT, a toolkit for training and deployment of neural Speech-to-Text models.

## 1 Introduction

Common Voice (Ardila et al., 2020) is a project aimed at collecting speech data for all of the world's languages. However, few of the released languages have been used to create publicly available Speech-to-Text (STT) models which are deployable by average developers. This report describes the process of creating developer-friendly models for the Common Voice languages with the least amount of data, using the Coqui STT toolkit.[1] Furthermore, we explore practical applications of the resulting models and release all models under a free/open licence.[2]

There are many dimensions along which one can judge machine learning models. For Speech-to-Text models, a common dimension is transcription accuracy, as measured by Word Error Rate and Character Error Rate. While this metric is important, we find that it should be considered in conjunction with other attributes of Speech-to-Text models, such as model size, latency, and time-to-deployment. All things being equal, *bigger models* are more accurate but they are expensive to train, and big models may not be deployable on most commonly found hardware (i.e. CPUs or microcomputers). *Faster models* (i.e. lower latency) are usually preferred, but there is a practical trade-off between speed and accuracy (e.g. wide (slow) decoding beams are more accurate than narrow (fast) ones). *Time-to-deployment* is a consideration that may be difficult to quantify, but often outweighs any other STT attribute in both academia and production settings. Time to deployment is the amount of time it takes an engineer to deploy a model into a production pipeline. We as co-authors have interests both in academia and production, and as such we consider all these dimensions to be important.

Lastly, we believe that speech technologies should be available to everyone, regardless of their native language. When working with under-resourced[3] languages, as researchers we often focus on the accuracy results to the detriment of making something of utility to the language communities we hope to serve. And while the error rates described in this report might not be state of the art (SOTA), there are plenty of *good applications for non-SOTA Speech-to-Text*.[4]

## 2 Languages and data

This technical report includes as many of the languages in the Common Voice 6.1 release with as little data as possible. That is, we started training models for languages with the least amount of data, and worked our way up the list for 3 weeks. Constructed languages were omitted, and we found experimentally that it was not possible to produce results for languages with under 30 minutes of training data.[5] By the end of our sprint, we successfully ran

---

[1] https://github.com/coqui-ai/stt
[2] The models described here are available under the GNU Affero GPL v3.0 at https://itml.cl.indiana.edu/models/ and at https://coqui.ai/models.
[3] Our definition of an *under-resourced* language is a language for which there exists little free/open speech data or speech tools. It is often the case that under-resourced languages have millions of speakers, but data and tools may be scarce for political and economic reasons.
[4] With all credit to Church and Hovy (1993).
[5] Languages omitted for having too little data were: Votic vot, Hindi hi, Assamese as, Abkhaz ab, Vietnamese vi,

experiments for those languages with more than 30 minutes but less than 12 hours of data.

The set of resulting languages can be found in Table 1. This table includes information about the training set: the amount of data in hours, the number of audio clips, the number of unique speakers, and the size of the output alphabet.

We used the standard `train/dev/test` splits from the release. Although these were optimised for large-data languages such as English, we wanted to make sure that our results were comparable to other work and future work.

These standard splits were created according to the following requirements:

- No two speakers in the same section
- No two transcripts in the same section
- No two recordings of the same transcript in the same section

This meant that there was no overlap in training, development and test sets between recordings, speakers or sentences. And this explains why, although for example the total amount of hours of data for Basque is around 85 hours, the amount of training data is a fraction of this.

## 2.1 Preprocessing

We used the `commonvoice-utils`[6] package to perform basic linguistic preprocessing of the data for each language, such as:

1. define an output alphabet
2. normalise Unicode encoding (e.g. for certain languages combining characters were used, for others precomposed characters)
3. replace mis-encoded characters in the training data (e.g. in Chuvash the Latin $ç$ U+00E7 misused in place of the Cyrillic $ҫ$ U+04AB
4. discard or replace punctuation on a per-language basis (e.g. keeping the apostrophe character if it is an orthographic symbol — as in Breton $c'h$ — or discarding it).

The package supports most of the languages in Common Voice and all the languages involved in these experiments.

---

and Punjabi `pa-IN`.
[6]https://github.com/ftyers/commonvoice-utils

## 3 Baseline models

The baseline training set up was based on Ardila et al. (2020). We used transfer learning from version `v0.9.3` of the English model[7] with the default hyperparameters and dropped the two final layers of the source model. The key training hyperparameters consisted of a dropout of $0.05$, a learning rate of $0.001$ and SpecAugment (Park et al., 2019) turned off. The batch size used was $8$, the maximum that could be fit on the GPU used for training.[8] We trained a baseline model for each language over 25 epochs, without early-stopping. The motivation behind using these hyperparameters was to get a comparison point before varying them on a per-language basis.

### 3.1 Results

The results from the baseline experiments are presented in Table 2 and in Figure 3a. We observe two general trends. The first trend is that the more data, the better the results. The language with the best results is Basque, which has the most seconds of data per alphabetic character.

A second trend is that the smaller the output alphabet, the better the results, even for the same quantity of data.

## 4 Parameter sweep

The first approach to improving the models was to use a hyperparameter sweep of a representative selection of languages. Taking the results from this subset of languages, we then re-train all languages based on the newly found best hyperparameter settings.

The motivation for this was that the default hyperparameter settings are unlikely to be optimal for all of the datasets. For example, smaller datasets are likely to converge better with a lower learning rate.

We selected six languages based on amount of data as our representative subset. In terms of low data, we chose Irish (`ga-IE`) and Odia (`or`). Both languages had around 30 minutes of training data. In addition they have either large alphabets or complex orthographic systems, making training more challenging. In terms of middle data we chose Chuvash (`cv`) and Breton (`br`). Chuvash has around 1 hour of training data, where Breton has around

---

[7]https://github.com/coqui-ai/STT/releases/tag/v0.9.3
[8]A single NVIDIA Tesla k40m with 12G of VRAM was used for all the experiments.

| Language | Autonym | Locale | Training | # Clips | # Speakers | $|V|$ |
|---|---|---|---|---|---|---|
| Irish | Gaeilge | `ga-IE` | 0:31:24 | 542 | 6 | 32 |
| Finnish | Suomi | `fi` | 0:32:29 | 456 | 1 | 28 |
| Odia | ଓଡ଼ିଆ | `or` | 0:32:56 | 389 | 1 | 82 |
| Hakha Chin | Hakha Chin | `cnh` | 0:38:14 | 807 | 6 | 27 |
| Romansch Vallader | Rumantsch | `rm-vallader` | 0:58:58 | 558 | 3 | 40 |
| Chuvash | Чӑвашла | `cv` | 1:04:44 | 932 | 2 | 39 |
| Lithuanian | Lietuvių kalba | `lt` | 1:10:17 | 928 | 2 | 36 |
| Upper Sorbian | Hornjoserbšćina | `hsb` | 1:26:58 | 799 | 1 | 42 |
| Sakha | Саха тыла | `sah` | 1:27:01 | 918 | 1 | 37 |
| Luganda | Luganda | `lg` | 1:36:09 | 1251 | 1 | 27 |
| Georgian | ქართული ენა | `ka` | 1:37:21 | 1055 | 3 | 34 |
| Turkish | Türkçe | `tr` | 1:52:31 | 1739 | 77 | 34 |
| Breton | Brezhoneg | `br` | 2:06:29 | 2781 | 5 | 37 |
| Romansch Sursilvan | Romontsch | `rm-sursilv` | 2:07:24 | 1381 | 7 | 36 |
| Indonesian | Bahasa indonesia | `id` | 2:09:49 | 2131 | 9 | 26 |
| Slovenian | Slovenščina | `sl` | 2:15:11 | 2038 | 2 | 29 |
| Latvian | Latviešu valoda | `lv` | 2:17:05 | 2553 | 2 | 35 |
| Tamil | தமிழ் | `ta` | 2:21:15 | 2010 | 7 | 49 |
| Maltese | Malti | `mt` | 2:31:23 | 2037 | 5 | 38 |
| Kyrgyz | Кыргызча | `ky` | 2:34:34 | 1955 | 3 | 37 |
| Greek | Ελληνικά | `el` | 2:45:04 | 2314 | 2 | 36 |
| Mongolian | Монгол хэл | `mn` | 3:02:38 | 2168 | 6 | 35 |
| Thai | ภาษาไทย | `th` | 3:24:57 | 2915 | 3 | 68 |
| Romanian | Românește | `ro` | 3:37:29 | 3369 | 2 | 32 |
| Dhivehi | ދިވެހި | `dv` | 3:56:12 | 2632 | 4 | 50 |
| Hungarian | Magyar nyelv | `hu` | 4:17:04 | 3339 | 2 | 36 |
| Estonian | Eesti | `et` | 5:00:16 | 2760 | 73 | 34 |
| Frisian | Frysk | `fy-NL` | 5:26:02 | 3923 | 6 | 34 |
| Portuguese | Português | `pt` | 7:40:45 | 6319 | 27 | 42 |
| Basque | Euskara | `eu` | 10:51:34 | 7505 | 53 | 28 |
| Tatar | Татарча | `tt` | 11:49:15 | 11181 | 2 | 40 |

Table 1: **Languages and data**. The list of languages for which models were trained. The list is ordered by ascending amount of training data. The amount of training data is shown in hours and number of clips. $|V|$ denotes number of symbols in the alphabet (dimensions of the softmax layer). The number of speakers is the number of unique speakers found in the training data. Locale codes are those provided by the Common Voice dataset release.

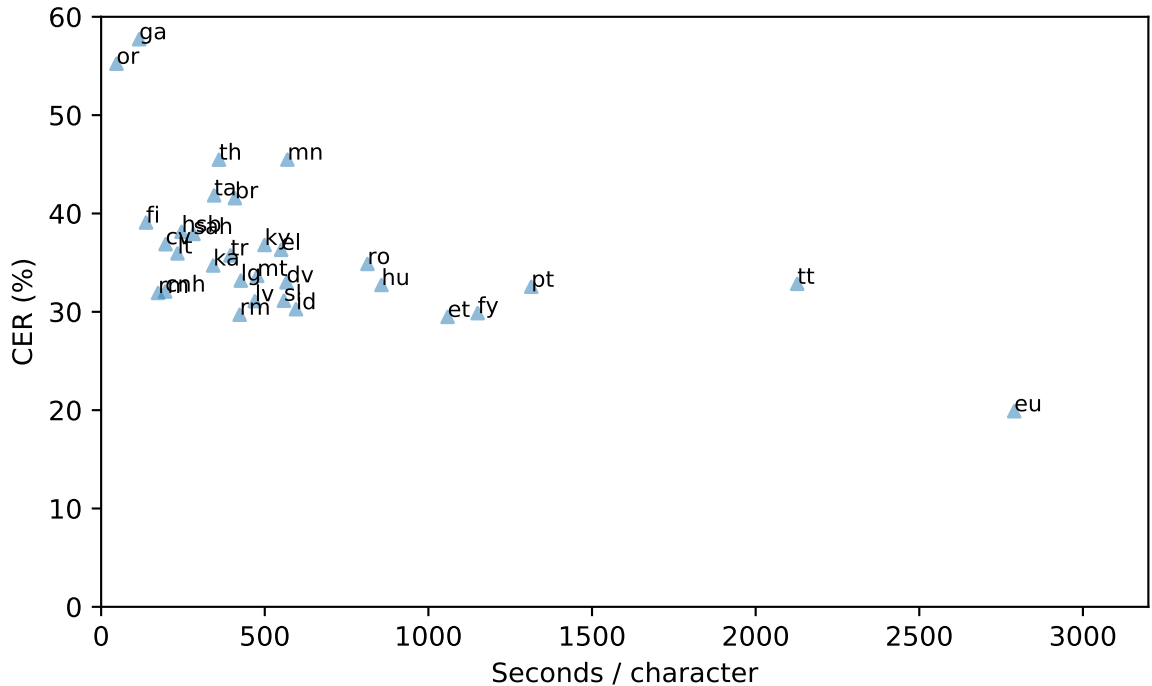

Figure 1: **Baseline results**. Character Error Rate (CER) for each language's baseline model plotted against the amount of training data per alphabetic character. There is a broadly linear relation between the amount of training data and CER.

2 hours. Both languages are among the worst performing for the amount of training data. Finally, for high data we chose Portuguese (`pt`) and Basque (`eu`). Basque is the best performing language and also has the most seconds of training data per character, while Portuguese is in the middle.

For the hyperparameter sweep we chose values for learning rate in the range $\{0.001, 0.0001, 0.00001\}$ and values for dropout in the range $\{0.2, 0.4, 0.6\}$. We also compared using SpecAugment or not. The configuration values we used for SpecAugment were:

- frequency_mask[p=0.8, n=2:4, size=2:4]
- time_mask[p=0.8, n=2:4, size=10:50, domain=spectrogram]

These were slightly more conservative versions of those mentioned in the original SpecAugment paper (Park et al., 2019).

The sweep resulted in total of 18 experiments per language, and each experiment lasted for ten epochs. The training and validation loss curves from these ten epochs were used to gauge the most promising hyperparameter settings.

We selected the hyperparameter combination where the validation loss was lowest while having the training loss higher on the tenth epoch. For example for Basque that was $[0.001, 0.2, \text{ON}]$, that is: *learning rate* of $0.001$, a *dropout* of $0.2$ and *SpecAugment* turned on.

Next, using these new found hyperparamters each language was trained over 100 epochs. We selected the best validation checkpoint over the entire 100 epochs, and tested this best validation model against the held-out test set.

### 4.1 Results

The results of model training after the parameter sweep are presented in Table 5 and in Figure 3b. As can be seen the improvements range from 2% to 30% with the majority being between 10%–15% (mean 14.25%, median 14.07%). The languages with the least data see the greatest improvement.

## 5 Addition of a language model

The Coqui STT inference engine[9] allows for the inclusion of an external language model to improve accuracy. This language model is an $n$-gram model based on KenLM (Heafield, 2011). It is often the case that text data is easier to acquire than audio data, and an external language model is a practical

---
[9] https://github.com/coqui-ai/STT

| Language | CER | WER | Loss |
| --- | --- | --- | --- |
| `ga-IE` | 57.72 | 94.30 | 65.12 |
| `fi` | 39.07 | 99.65 | 66.59 |
| `or` | 55.22 | 98.89 | 97.12 |
| `cnh` | 32.06 | 77.76 | 36.07 |
| `rm-vallader` | 31.92 | 92.02 | 69.97 |
| `cv` | 36.87 | 96.97 | 69.47 |
| `lt` | 35.94 | 98.81 | 73.12 |
| `hsb` | 38.13 | 96.24 | 90.38 |
| `sah` | 37.91 | 96.30 | 91.27 |
| `lg` | 33.17 | 97.67 | 66.52 |
| `ka` | 34.71 | 98.07 | 78.30 |
| `tr` | 35.73 | 95.32 | 54.26 |
| `br` | 41.56 | 94.87 | 41.36 |
| `rm-sursilv` | 29.71 | 89.17 | 55.90 |
| `id` | 30.27 | 89.67 | 41.45 |
| `sl` | 31.13 | 90.25 | 37.30 |
| `lv` | 31.08 | 88.27 | 32.63 |
| `ta` | 41.84 | 100.00 | 54.05 |
| `mt` | 33.65 | 93.65 | 61.05 |
| `ky` | 36.80 | 94.09 | 64.77 |
| `el` | 36.31 | 88.14 | 50.34 |
| `mn` | 45.48 | 96.72 | 107.70 |
| `th` | 45.47 | N/A | 55.70 |
| `ro` | 34.87 | 92.87 | 56.89 |
| `dv` | 33.00 | 94.73 | 76.81 |
| `hu` | 32.73 | 89.16 | 52.02 |
| `et` | 29.48 | 92.23 | 89.44 |
| `fy-NL` | 29.86 | 79.63 | 54.70 |
| `pt` | 32.55 | 84.10 | 53.52 |
| `eu` | 19.89 | 80.96 | 41.52 |
| `tt` | 32.85 | 90.99 | 44.63 |

Table 2: **Baseline results**. Character Error Rate (CER), Word Error Rate (WER), and CTC Loss reported on each language's held-out test set. The WER metric for Thai (`th`) does not apply because the language does not use spaces to delimit words.

| Language | Types | Tokens | Size |
| --- | --- | --- | --- |
| `ga-IE` | 324k | 28M | 79M |
| `fi` | 2.3M | 186M | 370M |
| `or` | 123k | 1.3M | 11M |
| `cnh` | 5.4k | 283k | 876K |
| `rm-vallader` | 58.3k | 1.1M | 6.1M |
| `cv` | 147k | 2.1M | 14M |
| `lt` | 196k | 2.9M | 18M |
| `hsb` | 49k | 325k | 4.0M |
| `sah` | 68k | 605k | 6.2M |
| `lg` | 88k | 1.6M | 9.1M |
| `ka` | 346k | 10.8M | 68M |
| `tr` | 2.1M | 654M | 1.2G |
| `br` | 225k | 5.8M | 33M |
| `rm-sursilv` | 57k | 1M | 5.8M |
| `id` | 390k | 98M | 193M |
| `sl` | 870k | 255M | 452M |
| `lv` | 159k | 2.9M | 16M |
| `ta` | 1M | 15M | 86M |
| `mt` | 591k | 42M | 125M |
| `ky` | 244k | 4.5M | 25M |
| `el` | 1.1M | 594k | 857M |
| `mn` | 28k | 249k | 2.3M |
| `th` | 6.3M | 12.5M | 82M |
| `ro` | 1.7M | 2G | 2.2G |
| `dv` | 76k | 419k | 6.8M |
| `hu` | 4.5M | 972M | 1.8G |
| `et` | 2.1M | 278M | 575M |
| `fy-NL` | 185k | 3.2M | 20M |
| `pt` | 2.4M | 1.4G | 1.9G |
| `eu` | 821k | 26.9M | 106M |
| `tt` | 201k | 4.5M | 25M |

Table 4: **Language modelling and text data**. Statistics on the generic $n$-gram language models trained for each language. Raw text sources include the Common Voice 6.1 train and development transcripts (TR), Wikipedia, OpenSubtitles and TED Talks. Size is the final size in bytes of the model on disk.

| Training data | Hyperparameters |
| --- | --- |
| Under 1h | $[0.00001, 0.2, \text{OFF}]$ |
| 1–10h | $[0.00001, 0.2, \text{ON}]$ |
| Greater than 10h | $[0.001, 0.2, \text{ON}]$ |

Table 3: **Sweep results**. Best hyperparameters found for different amounts of training data after running the sweep. The three values are *learning rate*, *dropout* and *SpecAugment*.

"low-hanging fruit" approach to take advantage of such text data in order to improve STT accuracy. In addition to the transcripts from the training and development set, we used corpora from the OPUS (Tiedemann, 2012) collection and from Wikipedia for training.

For the corpora from the OPUS collection we prioritised those which more closely represent speech, so for example OpenSubtitles and TED talks were prioritised if available. We excluded corpora representing interface localisation strings (Mozilla, Ubuntu, KDE etc.).

The corpora were processed using the `covo` tool from the `commonvoice-utils`[10] package. The processing involved: Sentence segmentation (only for Wikipedia), Unicode normalisation, and processing to match the output alphabet. All decoding hyperparameter settings were left as default.

For the variants of Romansh it was difficult to find freely-available corpora. In OPUS, the largest corpus (QED) was in fact Romanian not Romansh, although it was filed under the Romansh language code.[11] So we made do with the Romansh Wikipedia, this is written in *Rumantsch Grischun*, a standardised form of Romansh.

## 5.1 Results

We observe that the inclusion of a generic $n$-gram language model significantly improves transcription accuracy as measured by Word Error Rate. On average, we observe a 35.7% relative improvement in WER when using a language model during decoding of the test set. All results are reported in the leftmost columns of Table 5.

Some of the language models are very large, this was due to the lack of filtering. We anticipate that it would be possible to adjust the size of the model according to the usual size/accuracy trade offs. It is worth noting however that our best performing system, Basque (`eu`) had an impressive increase with a relatively small language model.

## 6 Effect of data size

To give an idea of the effect of training data size, we took our best performing model and trained it with increasing amounts of data. We started with 500 clips (0:43:18) and doubled the number of clips until we reached the total amount of data available. This resulted in five models.

## 6.1 Results

As can be seen in the Figure 2, the performance improves rapidly as we begin to double the amount of data, dropping off steeply towards the end. It is interesting to note that the addition of the language model does not drop off as steeply and we see a solid improvement of around 10% points both with 1/10 of the data and with the whole data.

## 7 Use cases

In this section we look at some practical use cases for the models we trained. The four tasks we look at are closed-vocabulary speech to text, keyword spotting, forced alignment, and pretranscription.

## 7.1 Closed-vocabulary Speech-to-Text

It is often the case that Speech-to-Text systems are used to recognise a small set of words, not transcribe generic speech. For example, in English-speaking countries it is common to call (via telephone) a bank, government office, or commercial entity and be greeted by a "voice-bot" which understands a small set of words. You might be asked to read aloud your credit card number, digit by digit. The system recognising your credit card number is probably a "closed-vocabulary" speech-to-text system. That is, the system is only intended to recognise numbers and will fail to transcribe anything else. This constraint is an intentional design feature. In addition to telephone-based voicebots, another common application for small-vocabulary STT models is in *command-and-control* settings. Command-and-control voice-powered applications range from coffee machines to hospital beds. These applications expect a small set of spoken instructions.

In narrowing the range of possible words an STT system may recognise, the system will perform better for the words of interest. For the under-resourced languages discussed here, open-domain Speech-to-Text may lead to undesirable accuracy levels, but these models may be very useful in a closed-vocabulary task.

### 7.1.1 Results

To test our models on a closed-vocabulary task, we used a small-vocabulary subset of the Common Voice corpus. The *Target Segments* sub-corpus includes the digits 0 through 9 and the words *yes* and *no* in 33 languages (see Table 8). The test set did not include any of the speakers from the training or development sets.

---

[10] https://github.com/ftyers/commonvoice-utils/

[11] We reported this issue upstream.

| Language | + Param. Sweep | | | + Language model | | | |
| --- | --- | --- | --- | --- | --- | --- | --- |
| | CER | Δ% | WER | CER | Δ% | WER | Δ% |
| ga-IE | 40.57 | -29.71 | 86.88 | 42.12 | -27.03 | 70.73 | -18.59 |
| fi | 30.69 | -21.45 | 96.65 | 27.92 | -28.54 | 60.54 | -37.36 |
| or | 35.00 | -36.62 | 95.00 | 36.05 | -34.72 | 74.58 | -21.49 |
| cnh | 26.48 | -17.40 | 67.36 | 24.65 | -23.12 | 53.28 | -20.91 |
| rm-vallader | 26.22 | -17.86 | 84.01 | 21.59 | -32.36 | 54.28 | -35.38 |
| cv | 33.73 | -8.53 | 95.37 | 33.10 | -10.23 | 64.98 | -31.87 |
| lt | 31.05 | -13.60 | 94.64 | 29.45 | -18.03 | 67.22 | -28.97 |
| hsb | 32.43 | -14.95 | 92.32 | 32.23 | -15.49 | 66.58 | -27.89 |
| sah | 36.33 | -4.18 | 94.50 | 39.57 | +4.37 | 72.00 | -23.82 |
| lg | 30.48 | -8.12 | 93.13 | 28.40 | -14.38 | 63.21 | -32.13 |
| ka | 31.13 | -10.32 | 95.75 | 28.09 | -19.08 | 59.83 | -37.51 |
| tr | 30.84 | -13.71 | 89.26 | 29.62 | -17.10 | 57.19 | -35.93 |
| br | 37.71 | -9.25 | 89.12 | 38.13 | -8.23 | 68.37 | -23.29 |
| rm-sursilv | 23.88 | -19.60 | 79.57 | 18.93 | -36.28 | 48.07 | -39.59 |
| id | 25.79 | -14.78 | 80.72 | 16.06 | -46.94 | 32.67 | -59.53 |
| sl | 26.79 | -13.95 | 82.36 | 18.16 | -41.65 | 40.33 | -51.03 |
| lv | 28.31 | -8.93 | 82.81 | 16.42 | -47.16 | 32.96 | -60.21 |
| ta | 46.58 | +11.32 | 99.93 | 49.36 | +17.97 | 100.00 | +0.07 |
| mt | 27.92 | -17.04 | 86.40 | 21.95 | -34.77 | 46.89 | -45.73 |
| ky | 30.55 | -16.98 | 87.07 | 26.33 | -28.45 | 52.19 | -40.06 |
| el | 31.20 | -14.07 | 80.21 | 24.35 | -32.92 | 48.84 | -39.12 |
| mn | 38.61 | -15.11 | 90.80 | 38.16 | -16.10 | 69.00 | -24.01 |
| th | 35.99 | -20.84 | 100.00 | 51.55 | +13.37 | 100.00 | -0.00 |
| ro | 28.00 | -19.70 | 82.12 | 18.55 | -46.81 | 36.34 | -55.75 |
| dv | 27.44 | -16.84 | 88.37 | 22.23 | -32.63 | 66.49 | -24.76 |
| hu | 31.00 | -5.28 | 85.87 | 22.28 | -31.94 | 44.27 | -48.44 |
| et | 24.99 | -15.25 | 85.53 | 19.62 | -33.47 | 46.05 | -46.16 |
| fy-NL | 26.49 | -11.29 | 74.05 | 19.77 | -33.81 | 41.20 | -44.35 |
| pt | 26.69 | -18.01 | 73.15 | 20.10 | -38.25 | 39.71 | -45.71 |
| eu | 15.65 | -21.33 | 68.69 | 6.99 | -64.87 | 20.64 | -69.95 |
| tt | 31.68 | -3.54 | 85.81 | 26.38 | -19.67 | 53.22 | -37.98 |

Table 5: **Results after parameter sweep and after adding a generic language model**. Character (CER) and Word Error Rate (WER) for each language when adding a language model.

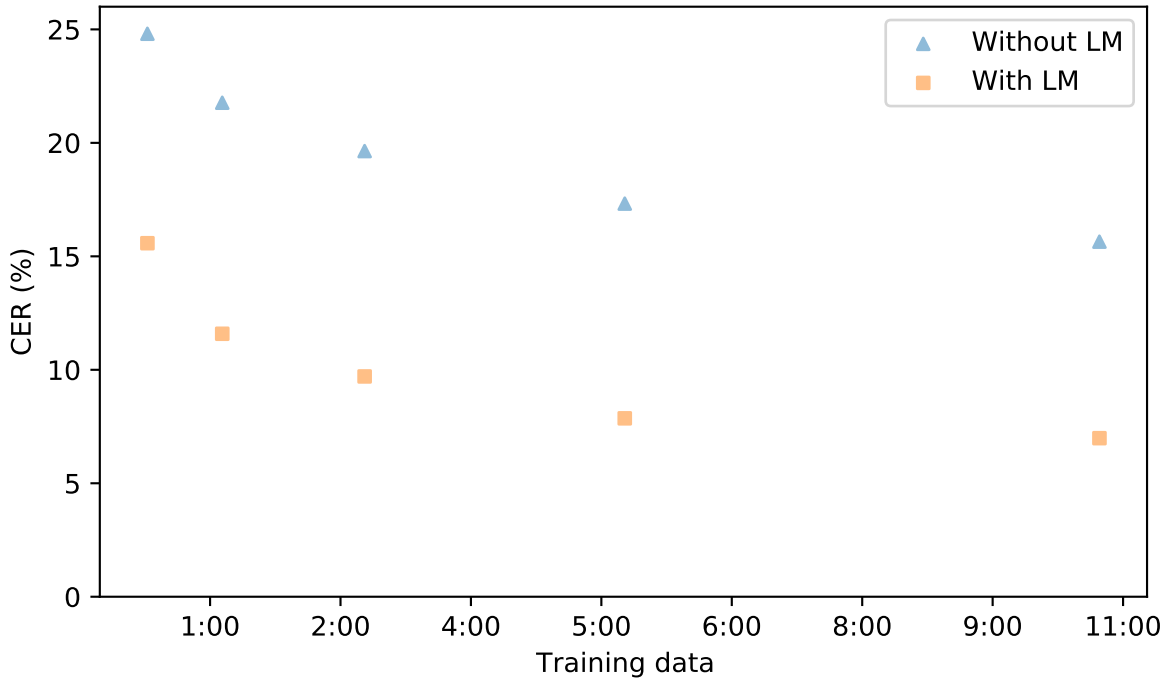

Figure 2: **Learning curve**. Character Error Rate (CER) for the best performing Basque model against amount of training data in hours. Results are reported both with and without a language model. The first data point is with 0:43:18 of training data and the last one is with 10:51:34. Each point approximately doubles the amount of training data.

The intersection of languages in the *Target Segments* corpus and languages in our experiments results in a total of 15 languages. We present results for these 15 languages in Table 6. Note that we did not train or fine-tune the core Speech-to-Text neural network for this experiment. We only created a small-vocabulary language model with *KenLM*, and report results on the *Target Segments* test sub-corpus.

We found that Basque produced the best results, with a WER of 3.9%, which may very well be deployable for certain applications. Other languages with less data experienced Word Error Rates closer to 20%. The task we created had a vocabulary of 12 words. Even smaller vocabularies (i.e. more constrained language models) may have better performance. The smallest practical vocabulary for deployment may be a binary *yes/no* recognition system, which may be practical for Interactive Voice Response systems (Perez-Marin, 2011).

### 7.2 Key-word spotting

Key-word spotting is the task of finding specific words in a given audio stream, often of continu-ous speech. This has a wide variety of uses, most notably *key-word search* and *wake-word detection*. Key-word search is when you have a large collection of audio saved on disk, and you want to identify all the instances of certain word. This is especially useful for information retrieval scenarios. Wake-word detection, on the other hand, is usually encountered in Internet of Things (IoT) applications such as smart speakers or connected devices. In wake-word detection, a continuous stream of audio is passed through a model, and if a key (wake) word is detected, the device performs some action. These two applications are similar in that they involve searching for a word in audio (i.e. not transcription), but these applications differ in terms of acceptable levels of accuracy. For wake-word detection, it is very important for the system to recognise all correct wake-words, and reject all non-wake-words. A false positive means the device wakes up when it shouldn't, and a false negative means the device ignores the user. For key-word search, there is in many cases more leniency, and thresholds for false positives and negatives is application specific.

| Language | Correct | Total | WER |
|---|---|---|---|
| cv | 22 | 26 | 15 |
| lg | 11 | 12 | 8 |
| ka | 13 | 17 | 23 |
| tr | 210 | 235 | 10.6 |
| br | 5 | 8 | 37 |
| id | 80 | 99 | 19 |
| sl | 80 | 99 | 19 |
| ta | 91 | 212 | 57.1 |
| ky | 4 | 5 | 20 |
| et | 4 | 5 | 20 |
| fy-NL | 64 | 77 | 16 |
| pt | 772 | 840 | 8.1 |
| eu | 196 | 204 | 3.9 |
| tt | 24 | 32 | 25 |

Table 6: **Small vocabulary results**. The results for the closed-vocabulary *target segments* corpus, where the language model consisted only of the words of interest. This consists of the digits 0 through 9, in addition to the words *yes* and *no*. Where the language did not have an exact translation for *yes* and *no*, a generic equivalent was used. We omit decimals for sample sizes below 100.

While there are specific algorithms for key-word spotting, cf. Mazumder et al. (2021), we use a very simple approach. We decode the audio as if we are performing a normal Speech-to-Text transcription task, and then we do a simple text search over the transcript.

For the experiments, we took the test set for each language, and selected 10 words at random from a set of those words longer than four characters to favour content words over function words. The results are presented in Table 7.

In general, precision is high and recall is medium to low. What does this mean in practice? Well, if you are planning to make something like a searchable interface to an audio archive, then a high precision means that if someone searches for a word, the results will likely contain that word. A low recall means that search results will not contain all instances of that word in the dataset.[12]

### 7.3 Forced alignment

Another possible use of the models is forced alignment. Forced alignment is useful to sync audio and text via generated timestamps, when you don't have the exact timestamps already. This is very useful

---

[12]An easy way to improve recall in a KWS application with Coqui STT is to use the `hotword boost` feature. This feature "boosts" the probability of a set of key words of interest.

| Language | Count | $F_1$ | Prec | Rec |
|---|---|---|---|---|
| `ga-IE` | 44 | 0.28 | 1.00 | 0.16 |
| `cv` | 122 | 0.24 | 1.00 | 0.14 |
| `sah` | 160 | 0.52 | 0.95 | 0.35 |
| `lg` | 180 | 0.32 | 0.94 | 0.19 |
| `tr` | 352 | 0.48 | 0.99 | 0.32 |
| `br` | 319 | 0.45 | 0.93 | 0.30 |
| `rm-surs.` | 286 | 0.54 | 0.96 | 0.38 |
| `id` | 840 | 0.54 | 0.98 | 0.37 |
| `lv` | 172 | 0.45 | 0.98 | 0.29 |
| `mt` | 303 | 0.44 | 0.93 | 0.28 |
| `ky` | 485 | 0.52 | 0.97 | 0.35 |
| `el` | 392 | 0.55 | 0.97 | 0.38 |
| `mn` | 696 | 0.51 | 0.91 | 0.36 |
| `ro` | 1008 | 0.47 | 0.87 | 0.33 |
| `hu` | 192 | 0.28 | 0.89 | 0.16 |
| `et` | 546 | 0.33 | 0.88 | 0.21 |
| `fy-NL` | 782 | 0.62 | 0.91 | 0.46 |
| `pt` | 1123 | 0.64 | 0.95 | 0.49 |
| `eu` | 1162 | 0.70 | 0.96 | 0.55 |
| `tt` | 833 | 0.50 | 0.96 | 0.34 |

Table 7: **Key-word spotting**. We show the dataset size, precision, recall and $F_1$ score for 20 languages. In general the precision is high and recall is moderate to low.

for aligning audio books to the original text of the book.

For this task we selected a collection of stories in the public domain Turkish book, *Çakıcı'nın İlk Kurşunu* by Sabahattin Ali. The audio of the book is available on YouTube read out by Melike Karadağ.[13]

We downloaded the audio of the book using `youtube-dl`[14] and extracted the text from an e-book PDF using `pdftotext` from *Poppler*.[15]

We used DSAlign[16] and our Turkish model to align the text with the extracted audio. It was used with its default hyperparameters. The Turkish model was 22nd out of 31 in terms of CER (see Table 5). DSAlign works as follows: It first segments the audio into utterances using `webrtcvad`, then uses the STT model to transcribe the utterances. Then uses the Smith–Waterman algorithm (Smith and Waterman, 1981) to align the utterances with segments in the text.

---

[13]https://www.youtube.com/watch?v=d9uJMcJtU9Y
[14]https://youtube-dl.org/
[15]https://poppler.freedesktop.org/
[16]https://github.com/mozilla/DSAlign/

There was 2:26:57 of audio and 21,218 tokens of text in the input files respectively. DSAlign generated alignments for 1,269 segments of audio out of a total of 2,806 totalling 0:53:46.

We tested the quality of the generated alignments by selecting a random collection of 100 and listening to them to verify if they matched the transcript. The success rate was 41/100. While this appears low, it still meant that we were able to extract 22 minutes of aligned audio with very little effort. Setting up DSAlign and preprocessing the data took less than 10 minutes.

### 7.4 Pretranscription

It may happen that you have some sound files that you would like to transcribe, but you do not want to transcribe the from scratch because of time constraints. In this case, you may be interested in transcribing using even fairly mediocre output from a speech recognition system as a rough first draft.

We used the Portuguese model (9th out of 31 in terms of CER) to perform two evaluations. In the first evaluation, a Spanish and Catalan speaker who was familiar with Portuguese (the first author) was asked to transcribe 100 clips of Portuguese randomly selected from the testset. For 50 of the clips he was given ASR output and for the other 50 he was given no ASR output.

We evaluated both reduction in CER compared to transcribing from scratch and speed compared to transcribing from scratch. The idea was to see if using the Portuguese model resulted in better transcripts produced more quickly than having a trained non-native speaker transcribe from scratch. The reader may ask here why evaluate this scenario? The answer being, while it is the case that for Portuguese it is quite easy to find native speakers who can do transcriptions, for a lot of other languages, much of the transcription is done by trained non-native speakers.

In the second evaluation, we looked at the time taken to postedit the text, given the original reference. This oracle simulates the best case of a native-speaker who can understand every utterance in one pass either typing it out or posteditting a draft transcription.

These two evaluation establish upper and lower bounds as to the use of ASR.

For the first evaluation, the annotator was able to achieve a CER of 59%. The CER of the ASR was 27%, and with postedition this was reduced to 17%. This makes a 42% point reduction in CER for an inexperienced annotator.

In terms of the second evaluation, we found that for the Portuguese system, the postedition speed with and without ASR was approximately equivalent. For 50 clips with ASR (1932 characters) it took 288 seconds to postedit them, leading to a characters/second of 6.7. For the 50 clips without ASR (2178 characters), it took 325 seconds leading to a character/second rate of 6.7.

It is worth noting that both quantitatively and qualitatively, posteditting the very bad ASR output took up most of the time, particularly making changes inside words, as opposed to splitting/joining or making changes at the edges, was time consuming. This could be improved by using a better interface that supported per-token prediction which the user could accept or not.

## 8 Discussion

In this report, we outline our approach to training new Speech-to-Text models for under-resourced languages. We show that for languages with as little as 30 minutes of data, transfer Learning allows us to create models which can be of practical use. It also shows that results can be improved substantially by judicious selection of hyperparameters and use of a language model.

We evaluated resulting models on the common tasks of closed-vocabulary STT, key-word spotting, forced alignment, and pretranscription. There are many more practical applications for models trained on small datasets, and we hope this work inspires more uses in more languages.

There are a number of avenues for improving the results. The default data splits for Common Voice exclude duplicate recordings of particular sentences. So if a sentence is recorded three times by different people, only a single recording will be left in the data. While this is potentially good practice for languages like English, where there is already a lot of data available, it is not ideal for languages with less data.

In addition it bakes in bias in terms of male voices being more represented in the dataset. So, we would like to include the additional audio clips in future work. We would also like to compare the effect of adding more speakers to adding more data from a single speaker. In several cases (most notably for Tatar, but cf. Table 9) we noticed that having a large amount of data from a single speaker caused the model to overfit and perform worse on the validation set.

We hope that this document will prove helpful to people intending to train and use speech recognition for their own languages, both in terms of providing a better starting point in terms of hyperparameters and in terms of outlining expected output quality and potential uses.

## Acknowledgements

Thanks to Alexandre Lissy for helping us with the experimental setup. We would also like to thank all the users who gave us encouraging feedback.

## A  Environmental impact

The upper bound on the total energy utilisation for these experiments is 635 MJ. We made this estimation by taking the approximate runtime (720 hours) and multiplying it by the thermal design power (TDP) of the GPU (245W for the NVIDIA Tesla k40m). Error could be introduced into these estimates from many sources: it ignores many other components of the machine; and the device is assumed to run at its TDP the entire runtime.

## B  Comparative graphs

Figures 3a, 3b and 3c show the results for the 31 languages plotted against the number of seconds training data per character in the alphabet. Figure 3a is a repeat of Figure 1 and included here for comparative purposes.

## C  Closed-vocabulary test set

In Table 8 we present the test set for the closed vocabulary experiments in Section 7.1.

## D  Dataset statistics

Table 9 presents statistics of validated hours for the languages we trained models for and how these are split into training, development and test sets.

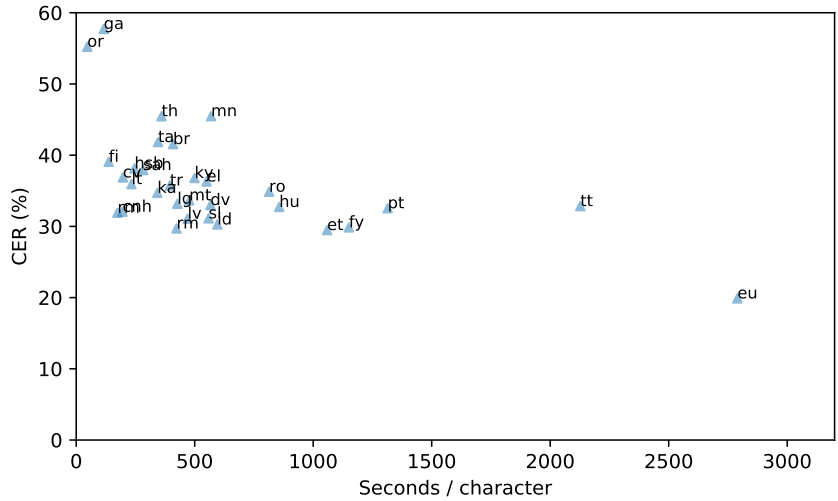

(a) **Baseline results**. Character Error Rate (CER) for each language's baseline model plotted against the amount of training data per alphabetic character. There is a broadly linear relation between the amount of training data and CER.

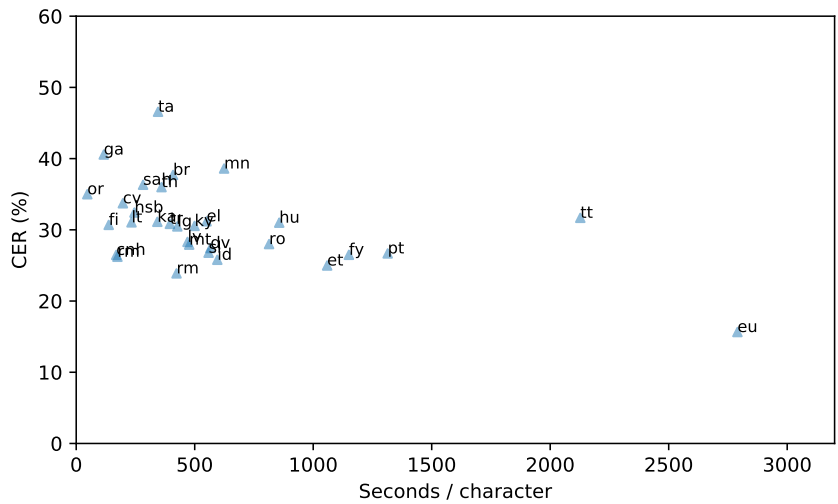

(b) **Results after parameter sweep**. Character error rate (CER) for post-parameter-sweep models against the amount of training data per alphabetic character. Adjusting the hyperparameters results in a 5-15% reduction in error rate.

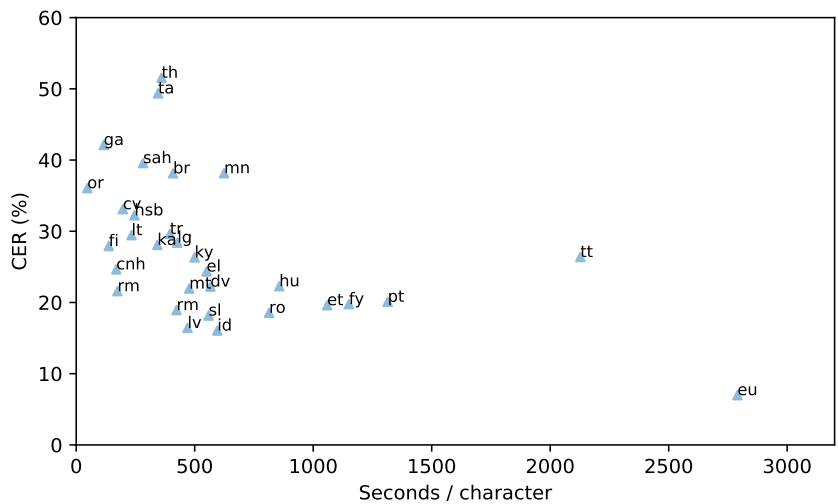

(c) **Results after adding a generic language model**. For most languages, the addition of a language model improves the results, and for Basque (eu) substantially improves them. The size of the language model seems to to be less correlated. For example, the Romanian (ro) language model at 2.2G is 20 times as big as the Basque one (106M), but the improvement is smaller.

|    | 0 | 1 | 2 | 3 | 4 | 5 | 6 | 7 | 8 | 9 | yes | no |
|----|---|---|---|---|---|---|---|---|---|---|-----|-----|
| cv | нуль | нèppe | иккĕ | виççĕ | тăваттă | пиллĕк | улттă | çиччĕ | саккăр | тăххăр | çапла | çук |
| lg | nooti | emu | bbiri | ssatu | nnya | ttaano | mukaaga | musanvu | munaana | mwenda | ye | nedda |
| ka | ნული | ერთი | ორი | სამი | ოთხი | ხუთი | ექვსი | შვიდი | რვა | ცხრა | დიახ | არა |
| tr | sıfır | bir | iki | üç | dört | beş | altı | yedi | sekiz | dokuz | evet | hayır |
| br | mann | unan | daou | tri | pevar | pemp | c'hwec'h | seizh | eizh | nav | ya | nann |
| id | nol | satu | dua | tiga | empat | lima | enam | tujuh | delapan | sembilan | ya | tidak |
| sl | nič | êna | dvé | trí | štíri | pét | šést | sédem | ósem | devét | ja | ne |
| ta | பூஜ்யம் | ஒன்று | இரண்டு | மூன்று | நான்கு | ஐந்து | ஆறு | ஏழு | எட்டு | ஒன்பது | ஆம் | இல்லை |
| ky | нөл | бир | эки | үч | төрт | беш | алты | жети | сегиз | тогуз | ооба | жок |
| et | null | üks | kaks | kolm | neli | viis | kuus | seitse | kaheksa | üheksa | jah | ei |
| fy-NL | nul | ien | twa | trije | fjouwer | fiif | seis | sân | acht | njoggen | ja | nee |
| pt | zero | um | dois | três | quatro | cinco | seis | sete | oito | nove | sim | não |
| eu | zero | bat | bi | hiru | lau | bost | sei | zazpi | zortzi | bederatzi | bai | ez |
| tt | ноль | бер | ике | өч | дүрт | биш | алты | җиде | сигез | тугыз | әйе | юк |

Table 8: **Closed-vocabulary test set.** The corpus used for the closed-vocabulary experiments.

| Locale | Validated | | Training | | | Development | | Test | |
|--------|-----------|---|----------|---|---|-------------|---|------|---|
|        | $|S|$ | Length | $|S|$ | Length | % | $|S|$ | Length | $|S|$ | Length |
| ga-IE | 83 | 2:43:26 | 6 | 0:27:06 | 16.58 | 16 | 0:26:29 | 61 | 0:29:53 |
| fi | 27 | 1:27:16 | 1 | 0:29:10 | 33.42 | 3 | 0:28:21 | 23 | 0:29:38 |
| or | 14 | 0:48:06 | 1 | 0:29:51 | 62.06 | 1 | 0:10:15 | 12 | 0:08:00 |
| cnh | 188 | 2:02:12 | 6 | 0:31:57 | 26.15 | 26 | 0:39:18 | 156 | 0:44:27 |
| rm-vallader | 32 | 2:05:05 | 3 | 0:56:21 | 45.05 | 4 | 0:32:15 | 25 | 0:35:46 |
| cv | 80 | 3:51:42 | 2 | 0:57:18 | 24.73 | 3 | 0:51:52 | 75 | 1:00:19 |
| lt | 22 | 2:00:42 | 2 | 1:03:41 | 52.76 | 2 | 0:19:35 | 18 | 0:37:12 |
| hsb | 18 | 2:19:15 | 1 | 1:22:26 | 59.20 | 1 | 0:16:48 | 16 | 0:40:01 |
| sah | 30 | 3:55:49 | 1 | 2:08:49 | 54.63 | 2 | 0:33:34 | 27 | 1:13:17 |
| lg | 57 | 2:48:25 | 1 | 1:26:24 | 51.30 | 2 | 0:31:14 | 54 | 0:50:44 |
| ka | 44 | 3:15:44 | 3 | 1:29:39 | 45.80 | 5 | 0:47:01 | 36 | 0:57:08 |
| tr | 631 | 17:29:56 | 77 | 1:44:02 | 9.91 | 128 | 1:39:28 | 401 | 1:50:38 |
| br | 124 | 5:46:07 | 5 | 1:43:59 | 30.04 | 12 | 1:20:12 | 107 | 1:34:30 |
| rm-sursilv | 75 | 5:11:35 | 7 | 1:57:08 | 37.59 | 17 | 1:37:34 | 51 | 1:36:53 |
| id | 176 | 8:13:36 | 9 | 1:53:04 | 22.91 | 20 | 1:42:05 | 141 | 1:52:04 |
| sl | 66 | 4:27:19 | 2 | 1:59:35 | 44.73 | 1 | 0:29:04 | 63 | 0:48:18 |
| lv | 88 | 5:15:48 | 2 | 1:57:06 | 37.08 | 6 | 1:37:01 | 80 | 1:41:15 |
| ta | 183 | 13:16:22 | 7 | 2:05:31 | 15.76 | 12 | 1:55:21 | 141 | 2:04:08 |
| mt | 147 | 6:41:42 | 5 | 2:15:09 | 33.64 | 18 | 1:43:31 | 124 | 2:03:43 |
| ky | 108 | 10:54:59 | 3 | 2:18:58 | 21.22 | 6 | 1:53:46 | 98 | 1:45:04 |
| el | 78 | 6:11:33 | 2 | 2:27:14 | 39.63 | 3 | 1:20:17 | 73 | 1:38:43 |
| mn | 299 | 10:11:50 | 6 | 2:46:54 | 27.28 | 31 | 2:34:24 | 262 | 2:43:11 |
| th | 143 | 7:26:56 | 3 | 3:02:08 | 40.75 | 9 | 1:54:09 | 131 | 2:30:36 |
| ro | 76 | 5:56:19 | 2 | 3:11:41 | 53.80 | 2 | 0:55:24 | 72 | 1:49:01 |
| dv | 155 | 16:30:35 | 4 | 3:42:16 | 22.44 | 9 | 2:56:57 | 142 | 2:56:08 |
| hu | 44 | 7:09:51 | 2 | 3:52:27 | 54.08 | 3 | 1:31:55 | 39 | 1:44:05 |
| et | 482 | 18:16:13 | 73 | 5:08:02 | 28.10 | 97 | 4:23:55 | 312 | 4:14:46 |
| fy-NL | 336 | 13:17:54 | 6 | 4:55:04 | 36.98 | 19 | 3:27:58 | 306 | 3:56:43 |
| pt | 840 | 45:05:39 | 27 | 7:02:48 | 15.63 | 100 | 5:03:30 | 648 | 5:33:02 |
| eu | 843 | 81:25:38 | 53 | 9:52:23 | 12.13 | 152 | 7:08:30 | 594 | 7:30:57 |
| tt | 164 | 22:42:28 | 2 | 10:21:39 | 45.63 | 2 | 1:49:27 | 159 | 4:00:01 |

Table 9: **Validated data and splits.** Statistics about the amount of validated data and the amount available in the training, development, and testing splits. $|S|$ is the number of unique speakers. The percentage (%) column shows the amount of validated data available to train with in the standard split. The numbers do not match exactly with those in Table 1 as they were calculated using different libraries and before/after preprocessing and filtering.